\pdfoutput=1

\documentclass[11pt]{article}

\usepackage{sec/acl}

\usepackage{times}
\usepackage{latexsym}

\usepackage[T1]{fontenc}

\usepackage[utf8]{inputenc}

\usepackage{microtype}

\usepackage{comment}

\usepackage{sec/package}
\usepackage{xcolor}
\definecolor{color1}{RGB}{102,194,165}
\definecolor{color2}{RGB}{141,160,203}
\definecolor{color3}{RGB}{252,141,98}
\definecolor{light-gray}{gray}{0.8}

\usepackage{todonotes}
\usepackage{inconsolata}

\title{
\textit{Original} or \textit{Translated}? 
A Causal Analysis of the Impact of Translationese
on 
Machine Translation Performance 
}

\author{Jingwei Ni\thanks{\hspace{0.1cm} Equal contributions.}
\\
  University College London \\
  \texttt{ucabjni@ucl.ac.uk} \\\And
  Zhijing Jin\samethanks \\
  MPI \& ETH Zürich \\
  \texttt{zjin@tue.mpg.de} \\\AND
  Markus Freitag \\
  Google Research \\
  \texttt{freitag@google.com} \\\And
  Mrinmaya Sachan \\
  ETH Zürich \\
  \texttt{msachan@ethz.ch} \\\And
  Bernhard Sch\"olkopf \\
  MPI \& ETH Zürich \\
  \texttt{bs@tue.mpg.de}
\\}

\begin{document}
\maketitle
\begin{abstract}
Human-translated text
displays distinct features from naturally written text in the same language.
This phenomena, known as \textit{translationese}, 
has been argued to confound the machine translation (MT) evaluation.
Yet, we find that existing work on translationese neglects some important factors and the conclusions are mostly correlational but not causal.
In this work, we collect \causalmt{}, a dataset where the MT training data are also labeled with the human translation directions. We inspect two additional critical factors, the \textit{train-test direction match} (whether the human translation directions in the training and test sets are aligned), and \textit{data-model direction match} (whether the model learns in the same direction as the human translation direction in the dataset). We show that these two factors have a large causal effect on the MT performance, in addition to the \textit{test-model direction mismatch} highlighted by existing work on translationese.
In light of our findings, we provide a set of suggestions for MT training and evaluation.\footnote{Our code and data are at \url{https://github.com/EdisonNi-hku/CausalMT}.}
\end{abstract}

\section{Introduction}

\begin{figure}[t]
    \centering
    \includegraphics[width=\columnwidth]{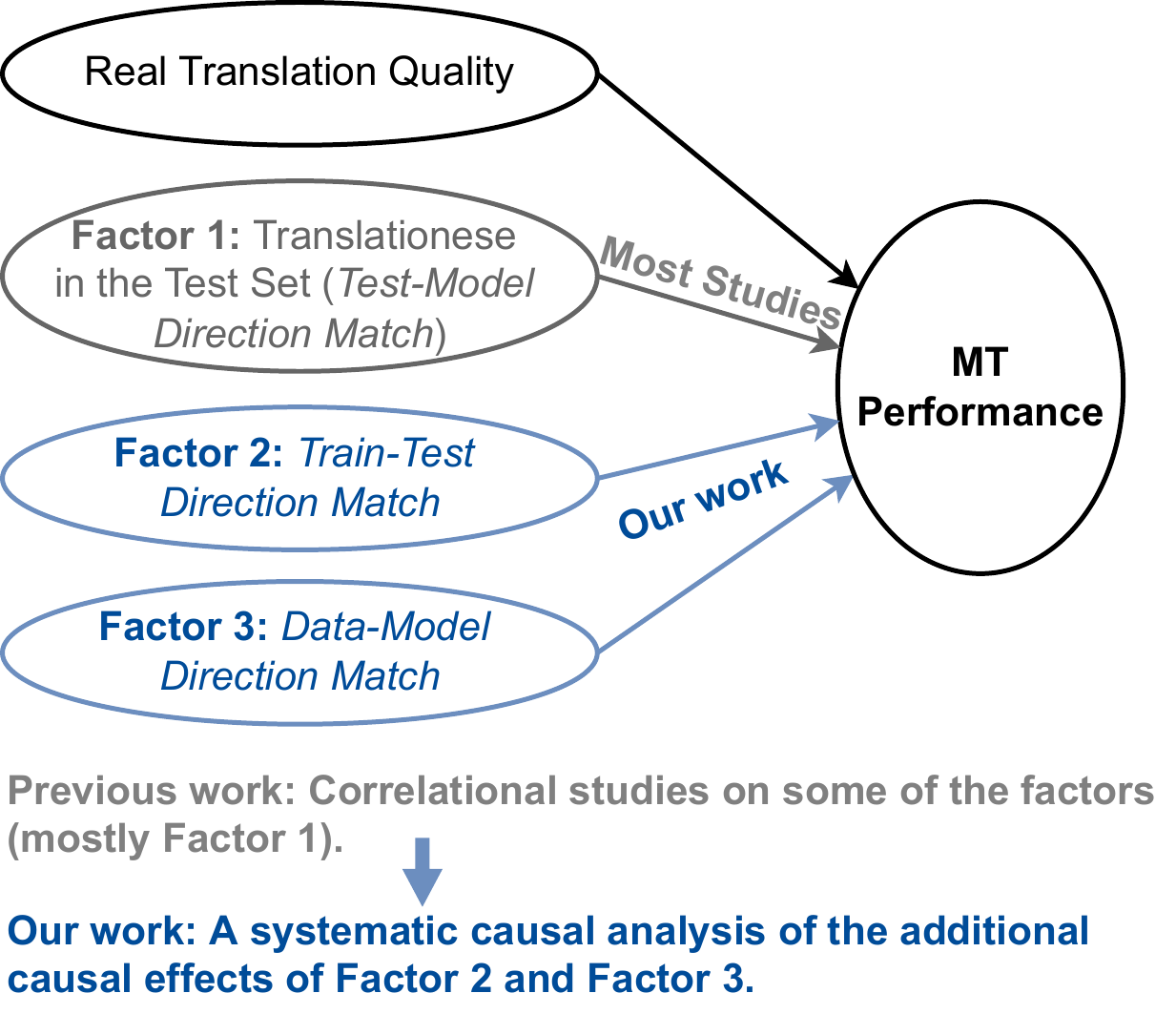}
    \caption{Three different factors illustrate the impact of translationese on MT performance. Previous work mainly focuses on how translationese in the test set (Factor 1) inflates BLEU score and makes it favor some translation systems over others. Our work investigates the causal effects of the other two key factors, the train-test direction match (Factor 2; i.e., whether the training set and the test set share the same human translation direction), and data-model direction match (Factor 3; i.e., whether the dataset collection direction and model translation direction are the same).
    }
    \vspace{-0.7em}
    \label{fig:causal_graph_overall}
\end{figure}
MT has long been concerned with the artifacts introduced by \textit{translationese}, the human-translated text that is systematically different from naturally written text in the same language~\cite{toury1980search,gellerstam1986translationese,toury1995descriptive,baker1993corpus,baroni2006new}. 
For a translation system translating from language $X$ to language $Y$, there can be two types of test data: sentences that originated in language $X$ and are human-translated into language $Y$ (denoted as $X$\humantrans $Y$), and sentences that originated in language $Y$ and human-translated into language $X$ (denoted as $X$\revhumantrans $Y$).\footnote{Note that the scope of this work does not cover pivot translation through a third language, but we encourage exploration in future work.} The main concern raised by this distinction of the two sets is whether the reported performance on a mixed test set truly reflects the actual translation quality. Previous work in MT has shown that translationese is a confounder in evaluating translation quality~\cite{lembersky-etal-2012-language,toral-etal-2018-attaining,laubli-etal-2018-machine,freitag2020bleu}.

Recent studies on causality have also brought to attention the importance of distinguishing the \textit{data-model direction match}, namely whether the {data collection direction} is the same as or opposite to the {model direction}, also known as \textit{causal} or \textit{anti-causal} learning~\cite{jin-etal-2021-causal,veitch2021counterfactual,scholkopf2012causal}. If the dataset is collected by human annotators who see the input $X$ and produce an output $Y$, then learning an \forwardmodel{} model is causal learning, and learning a \backwardmodel{} model is anticausal learning.

In this work, we 
study 
the artifacts in MT brought by translationese from the viewpoint of causality, specifically, the match between the data and model directions.
We consider two factors of variation in MT: \textit{human translation direction} (in
both the training and the test set)
and \textit{model translation direction}.
Thus, we formulate the effect of translationese in the test set as the \textit{test-model direction match} problem, 
and causal/anticausal learning as the \textit{data-model direction match} problem. Further, we identify the third important factor, the \textit{train-test direction match} problem, namely, whether the training set and the test set are collected with the same human translation direction.
We study the causal influences of these three factors
on MT performance in Figure~\ref{fig:causal_graph_overall}.
Previous work has mainly studied the artifacts introduced by the test-model direction match on MT performance~\cite{toral-etal-2018-attaining,graham-etal-2020-statistical,edunov-etal-2020-evaluation}, but little considers the other two factors, the train-test and data-model direction match \cite{kurokawa-etal-2009-automatic,larkin-etal-2021-like}. Moreover, most analyses are based on correlation but not causation
\cite{graham-etal-2020-statistical}.

In this paper, we use causal inference~\cite{pearl2009causality,peters2017elements} to re-investigate previous conclusions about the most studied relationship between the test-model direction match and MT performance. We calibrate the effect of test-model direction match by analyzing the causal effects of the two other factors, the train-test and data-model direction match. 
In our causal analyses, we use interventions to study the effect of train-test direction match, and covariate matching to study the effect of data-model direction match, by controlling for covariates such as sentence lengths and content.

We build \causalmt{}, a new dataset on five language pairs labeled with the human translation directions, and statistically verify that translationese tends to be simpler and more verbose, corroborating previous observations on translationese
\cite{toury1980search,gellerstam1986translationese,toury1995descriptive,baker1993corpus}.
Then, we rigorously analyze \causalmt{}, leading to the following new insights and contributions:
\begin{enumerate}[itemsep=0pt,topsep=1pt,label=C\arabic*.]
    \item Previous work claims that translationese in the test set inflates MT model performance and thus suggests removing the translationese-to-original half of the test set~\cite{toral-etal-2018-attaining,zhang-toral-2019-effect,graham-etal-2020-statistical,barrault-etal-2019-findings}.
    Our work shows that the translationese-to-original half of the test set does not necessarily inflate MT performance in all cases. In some cases, it can even be more challenging than the other half, depending on the human translation direction in the training corpus. Hence, we suggest still reporting performance on both test sets, and also reporting the training data direction if available.
    \item Previous work~\cite{burlot-yvon-2018-using} claims that back-translation (BT)~\cite{sennrich-etal-2016-improving} is usually more effective than self-training (ST)~\cite{he2019selftraining}.
    Our work shows that BT is not necessarily better than ST in all cases. This result also depends on how the pseudo-parallel corpus aligns with the human translation direction in the test set. We suggest choosing BT or ST depending on this direction match.
    \item Previous work claims that BT's performance improvement is largely reflected on the translationese-to-original half of the test set, but the improvement is very small on the other half~\cite{toral-etal-2018-attaining,freitag-etal-2019-ape}.
    Our work shows that the improvement of BT can be larger on the other half of the test set as well, as long as the pseudo-parallel corpus aligns with the human translation direction in the test set.
    \item 
    Our work shows that
    data-model direction match also has a large causal effect on the MT performance of up to 12.25 BLEU points after adjusting for other covariates using the backdoor adjustment~\cite{pearl1995causal}.
\end{enumerate}



\section{\causalmt{} Dataset}
\begin{table*}[t]
    \centering
    \small
    \setlength\tabcolsep{2pt}
    \resizebox{\textwidth}{!}{
    \begin{tabular}{lcc|cc|cc|cc|cc}
    
    \toprule
    & 
    De\humantrans{}En & De\revhumantrans{}En & De\humantrans{}Fr & De\revhumantrans{}Fr & En\humantrans{}Fr & En\revhumantrans{}Fr & En\humantrans{}Es & En\revhumantrans{}Es & Es\humantrans{}Fr & Es\revhumantrans{}Fr \\ \hline
    \# Training Samples & 248K & 248K & 220K & 220K & 203K & 203K & 93K & 93K & 92K & 92K \\
    \# Words/Sample & 22.4/25.5 & 22.9/23.9 & 22.6/28.7 & 25.4/30.4 & 24.5/28.9 & 27.5/30.5 & 24.0/25.7 & 31.6/31.9 & 32.4/36.5 & 27.9/30.5 \\
     \# Sents/Sample & 1.05/1.04 & 1.04/1.02 & 1.05/1.86 & 1.07/1.94 & 1.03/1.89 & 1.05/1.95 & 1.03/1.05 & 1.08/1.08 & 1.09/2.18 & 1.07/1.95 \\
     Passive Voice (\%) & -/12.90 & -/11.48 & -/- & -/- & 11.70/- & 13.45/- & 11.49/- & 14.94/- & -/- & -/- \\
    {Vocab Size} & 119K/37K & 113K/40K & 108K/49K & 106K/55K & 40K/53K & 38K/56K & 29K/46K & 26K/47K & 47K/38K & 48K/42K \\
    {Expansion Factor} & en:de=1.13 & en:de=1.04 & fr:de=1.26 & fr:de=1.19 & fr:en=1.18 & fr:en=1.10 & es:en=1.06 & es:en=1.01 & fr:es=1.12 & fr:es=1.09 \\
    \bottomrule
    \end{tabular}
    }
    \caption{
    Detailed characteristics of the \causalmt{} dataset. We first report the number of translation pairs in the training set (\# Training Samples), and for each parallel corpus ($X$\humantrans{}/\revhumantrans{} $Y$), we report the following statistics  in both language $X$ and $Y$ (denoted as ``\textit{stats in $X$/stats in $Y$}''): the number of words per sample
    (\# Words/Sample),
    number of sentences per sample
    (\# Sents/Sample),
    percentage of samples with passive voice, vocabulary size, and the expansion factor. The expansion factor from language $X$ to language $Y$ ($X$:$Y$) is calculated by the average word count per sample in language $X$ divided by the average word count per sample in language $Y$.
    }
    \vspace{-0.7em}
    \label{tab:corpus_complexity}
\end{table*}


To investigate the effect of train-test direction match and data-model direction match, we need to collect translation data in different human translation directions.

\subsection{Data Collection}

To construct our \causalmt{} dataset consisting of a large number of translation pairs labeled with the human translation direction,\footnote{Most existing datasets do not distinguish the human translation direction for the training set~\cite{kolias2014exploratory, barrault-etal-2019-findings}. Some works train a classifier to identify the human translation direction ~\cite{kurokawa-etal-2009-automatic,riley-etal-2020-translationese}, but they are not our ideal choice since this classification may rely on the domain difference of the two directions~\cite{rabinovich-wintner-2015-unsupervised}. Our dataset is an extended version of our prior study \cite{jin-etal-2021-causal}, but ours is significantly larger to enable the various analyses in our study.} we use the EuroparlExtract toolkit~\cite{ustaszewski2019optimising} to filter translation pairs by meta-information (e.g., the tag specifying the original language of the speaker). Specifically, in the Europarl corpus~\cite{koehn-2005-europarl}, we iterate over each transcript that has an origination label and mark a sentence as original text if the original language of the speaker is the same as the language this sentence is in, or otherwise mark it as the translated text.
After extracting the direction-labeled language pairs, we remove all duplicates in the entire dataset. 
Since our study needs to compare training on parallel corpora of the same language pair but with two different human translation directions, 
e.g., De\humantrans{}En and De\revhumantrans{}En,
we control the size of the two corpora to be the same by downsampling the larger set. 

Among all language pairs we can obtain, we keep five language pairs with the largest number of data samples. 
As in Table~\ref{tab:corpus_complexity}, the \causalmt{} dataset contains over 200K translation pairs in each training set
of three language pairs and over 90K translation pairs in each training set of the other two language pairs. The development set and test set contain 1K and 2K translation pairs for all language pairs in each direction, respectively.

\subsection{Dataset Characteristics}

We analyze the characteristics of the \causalmt{} dataset in light of how translated text differs from naturally written text in the same language.


Our findings echo the observations by previous work on the distinct features of translationese
\cite{toury1980search,gellerstam1986translationese,toury1995descriptive,baker1993corpus,baroni2006new,volansky2015features}.
For example, translationese tends to be simpler and more standardized~\cite{baker1993corpus,toury1995descriptive,laviosa1998universals}, such as having a smaller vocabulary and using certain discourse markers more often~\cite{baker1993corpus,baker1995corpora,baker1996corpus}. Translationese also tends to be influenced by the source language in terms of its lexical and word order choice~\cite{gellerstam1986translationese}.

In the \causalmt{} data, we observe three properties. 
(1) Within each language pair (e.g., German and English), the same language's \textit{translationese always has a smaller vocabulary} than its naturally written text corpus. For example, the translationese German in De\revhumantrans En has only 113K vocabulary, which is 5K smaller than the vocabulary of the German corpus in De\humantrans En.
(2) \textit{Translationese tends to be more verbose}. For each language pair, we calculate the expansion factor from language $X$ to language $Y$ ($X$:$Y$) as the average word count per sample in language $X$ divided by the average word count per sample in language $Y$. For example, for each (English, German) translation pair, the number of English words is 1.13 times that of German words when English is the translationese
(i.e., en:de expansion factor=1.13). On the other hand, the en:de expansion factor is only 1.04 when English is the naturally written text.
(3) We use a syntax-based parser to detect the percentage of samples with passive voice in English, details of which are in \cref{appd:linguistic}.
There is a clear distinction that \textit{translationese English tends to use more passive voice than original English}, e.g., 14.94\% translationese samples in the passive voice in the En\revhumantrans{}Es corpus in contrast with 11.49\% original English samples in the reverse direction.

\vspace{-1pt}

\section{The Overshadowing Effect of Train-Test Direction Match}
\vspace{-1pt}

The first analysis of this paper aims to calibrate the most studied relationship of the {test-model direction match} and MT performance by considering
the additional effect of the \textit{train-test direction match}.

\myparagraph{Previous work observes that the translationese-to-original test set inflates the score.}
To evaluate a model with the $X$-to-$Y$ translation direction, traditionally, the test set is a mixture of two halves, one with the human translation direction $X$\humantrans$Y$ (aligned) and the other $X$\revhumantrans$Y$ (unaligned, or translationese-to-original)~\cite{bojar-etal-2018-findings}.

Previous studies propose that the unaligned, translationese-to-original test set is easier to translate than the other aligned test set because translationese inputs are easy for the MT model to handle~\cite{toral-etal-2018-attaining,zhang-toral-2019-effect,graham-etal-2020-statistical}. The inflated test performance caused by translationese has long been speculated 
\cite{lembersky-etal-2012-language,toral-etal-2018-attaining,laubli-etal-2018-machine}, and recent work has statistically verified the correlation~\cite{graham-etal-2020-statistical}.

With the previous understanding, some works suggest removing the unaligned half of the test set~\cite{toral-etal-2018-attaining,zhang-toral-2019-effect,graham-etal-2020-statistical}, which was adopted by the 2019 WMT shared task~\cite{barrault-etal-2019-findings}, whereas others suggest keeping both but reporting the performance separately~\cite{freitag-etal-2019-ape,edunov-etal-2020-evaluation}. The motivations from the two sides are that in the unaligned half, although the source text being translationese is an easy input to the model, its target text being naturally written text makes the evaluation more natural.

\myparagraph{This ``inflation'' depends on train-test direction match.}
We take a step back from the argument on whether the unaligned test set positively or negatively affects the MT performance evaluation. Instead, we 
call attention to the fact that, beyond the test-model direction match, there can be other factors also playing a critical in the MT performance evaluation, i.e., 
the train-test direction match.
\begin{table}[t]
    \centering
    \small
    \setlength\tabcolsep{0pt}
    \setlength\extrarowheight{-9pt}
    \begin{tabular}{cc|c|cc|ccc}
    \toprule
    \multicolumn{3}{c|}{\textbf{De-to-En Translation}} & \multicolumn{3}{c}{\textbf{En-to-De Translation}}  \\
    $\alpha\%$ & T1 (de, en$^*$) & T2 (de$^*$, en) & $\alpha\%$ & T1 (en, de$^*$) & T2 (en$^*$, de) \\\hline
0\% & \bluebox{24.68} & \redbox{35.86} & 0\% & \bluebox{21.24} & \redbox{26.27} \\
25\% & \bluebox{28.98} & \redbox{35.40} & 25\% & \bluebox{25.60} & \redbox{25.44} \\
50\% & \bluebox{30.86} & \redbox{34.53} & 50\% & \bluebox{27.29} & \redbox{24.70} \\
75\% & \bluebox{31.52} & \redbox{31.92} & 75\% & \bluebox{27.82} & \redbox{23.23} \\
100\% & \bluebox{31.33} & \redbox{27.07} & 100\% & \bluebox{28.94} & \redbox{20.32} \\

    \bottomrule
    \multicolumn{3}{c|}{\textbf{De-to-Fr Translation}} & \multicolumn{3}{c}{\textbf{Fr-to-De Translation}}  \\
    $\alpha\%$ & T1 (de, fr$^*$) & T2 (de$^*$, fr) & $\alpha\%$ & T1 (fr, de$^*$) & T2 (fr$^*$, de) \\\hline
0\% & \bluebox{24.37} & \redbox{36.44} & 0\% & \bluebox{18.85} & \redbox{22.62} \\
25\% & \bluebox{28.60} & \redbox{36.21} & 25\% & \bluebox{24.30} & \redbox{22.88} \\
50\% & \bluebox{28.87} & \redbox{34.06} & 50\% & \bluebox{25.91} & \redbox{22.10} \\
75\% & \bluebox{30.11} & \redbox{32.42} & 75\% & \bluebox{27.41} & \redbox{20.94} \\
100\% & \bluebox{30.45} & \redbox{27.65} & 100\% & \bluebox{27.79} & \redbox{18.68} \\

    \bottomrule
    \multicolumn{3}{c|}{\textbf{En-to-Fr Translation}} & \multicolumn{3}{c}{\textbf{Fr-to-En Translation}}  \\
    $\alpha\%$ & T1 (en, fr$^*$) & T2 (en$^*$, fr) & $\alpha\%$ & T1 (fr, en$^*$) & T2 (fr$^*$, en) \\\hline
0\% & \bluebox{31.74} & \redbox{38.09} & 0\% & \bluebox{31.91} & \redbox{40.74} \\
25\% & \bluebox{36.64} & \redbox{37.84} & 25\% & \bluebox{35.94} & \redbox{38.69} \\
50\% & \bluebox{38.00} & \redbox{36.83} & 50\% & \bluebox{37.36} & \redbox{37.51} \\
75\% & \bluebox{39.00} & \redbox{36.10} & 75\% & \bluebox{39.11} & \redbox{36.61} \\
100\% & \bluebox{39.74} & \redbox{33.88} & 100\% & \bluebox{40.27} & \redbox{33.01} \\

    \bottomrule
    \multicolumn{3}{c|}{\textbf{En-to-Es Translation}} & \multicolumn{3}{c}{\textbf{Es-to-En Translation}}  \\
    $\alpha\%$ & T1 (en, es$^*$) & T2 (en$^*$, es) & $\alpha\%$ & T1 (es, en$^*$) & T2 (es$^*$, en) \\\hline
0\% & \bluebox{31.74} & \redbox{38.09} & 0\% & \bluebox{31.91} & \redbox{40.74} \\
25\% & \bluebox{36.64} & \redbox{37.84} & 25\% & \bluebox{35.94} & \redbox{38.69} \\
50\% & \bluebox{38.00} & \redbox{36.83} & 50\% & \bluebox{37.36} & \redbox{37.51} \\
75\% & \bluebox{39.00} & \redbox{36.10} & 75\% & \bluebox{39.11} & \redbox{36.61} \\
100\% & \bluebox{39.74} & \redbox{33.88} & 100\% & \bluebox{40.27} & \redbox{33.01} \\

    \bottomrule
    \multicolumn{3}{c|}{\textbf{Es-to-Fr Translation}} & \multicolumn{3}{c}{\textbf{Fr-to-Es Translation}}  \\
    $\alpha\%$ & T1 (es, fr$^*$) & T2 (es$^*$, fr) & $\alpha\%$ & T1 (fr, es$^*$) & T2 (fr$^*$, es) \\\hline
0\% & \bluebox{37.32} & \redbox{46.25} & 0\% & \bluebox{39.16} & \redbox{41.60} \\
25\% & \bluebox{40.60} & \redbox{46.43} & 25\% & \bluebox{41.81} & \redbox{40.64} \\
50\% & \bluebox{41.94} & \redbox{45.57} & 50\% & \bluebox{43.48} & \redbox{39.66} \\
75\% & \bluebox{42.39} & \redbox{43.88} & 75\% & \bluebox{45.13} & \redbox{39.03} \\
100\% & \bluebox{42.46} & \redbox{40.00} & 100\% & \bluebox{45.42} & \redbox{37.56} \\
    \bottomrule
    \end{tabular}
    \caption{BLEU points of all five language pairs on training sets mixed by $\alpha\%$ $X$\humantrans{}$Y$ and $(1-\alpha\%)$ $X$\revhumantrans{}$Y$ data, where the mixture rate $\alpha=0,25,50,75,100$. We always use T1 to denote the test set aligned with the model direction, and T2 to denote the unaligned one. For readability, we use $^*$ to denote the translationese language. For example, ``(de, en$^*$)'' means original German and translated English pairs.}
    \vspace{-0.7em}
    \label{tab:consis_train_test}
\end{table}

For a given machine translation task to learn the \forwardmodel{} translation, there can be two questions: the question from previous work is whether we should use the test set aligned with the model translation direction (T1) or the test set unaligned with the model translation direction (T2) to evaluate the model fairly, whereas the question answered by our work is \textit{which training data should be used to achieve the best performance}.


Our analysis aims to obtain causal conclusions on how intervening on the train-test direction match affects the MT performance. Therefore, we control all other possible confounders. For each language pair, we control the total training data size to be the same\footnote{A side benefit of controlling the training data size is that our experiments can help answer what the best nature (i.e., human translation direction) of the training data given a fixed annotation or computation budget is. We leave the space for future work to increase the total training set size with all available training data in both directions.} when varying the portion of data in two directions. We also enumerate all other possible interventions, such as varying the two model translation directions and reporting performance on two different halves of the test set with two human translation directions. 
We also control that all translation models use the same
Transformer architecture~\cite{vaswani2017attention} by fairseq~\cite{ott-etal-2019-fairseq}, with experimental details in Appendix~\ref{appd:implementation_details}.
We report the experiment results of how intervening the train-test direction match affects the MT performance in BLEU score~\cite{papineni-etal-2002-bleu} in Table~\ref{tab:consis_train_test}.
The main takeaways are as follows:

(1) It is not always the case that, for the same model, the unaligned test set T2 yields higher/more inflated results than the aligned test set T1. 
When the training data has 75--100\% aligned training samples, performance reported on T2 is, in most cases, no longer larger than that on the other half. With such training data, usually, T1 inflates the BLEU score more.

(2) 
The train-test direction match can have an overshadowing effect over the artifacts introduced by the translationese-to-original test set, since no matter which test set we use, the more matched the train and test directions are, the higher the performance reported on T1 than T2 is.
Specifically, as we vary the portion of the aligned training data from 0 to 100\%, the performance on T1 keeps increasing, and the performance on T2 keeps decreasing.
Additionally, if the training data has about 0--50\% samples aligned with the model translation direction, then, in many cases, T2 is higher than T1, which might explain the previous observations that T2 inflates the BLEU score~\cite{toral-etal-2018-attaining,graham-etal-2020-statistical}.
To account for another possible interpretation, such as the domain shift between the training and test sets, we also conduct an additional evaluation using the newstest2014 test sets, which do not share any domain similarity with our training sets, but still support our observation (in Appendix Table~\ref{tab:consis_train_test_newstest}).



Hence, the two constructive suggestions for future work are that (1) it is important to still report on both test sets, and also the training data direction if available, and (2) parallel training data in the same direction with the test set is more helpful.

\myparagraph{Monolingual data in the original language of the test set is more helpful.}
With the intuition that the train-test direction match is a crucial factor for MT performance, we also look into its implications on semi-supervised learning.

Given additional monolingual data, a common question in MT is what type of monolingual data to use, and the accompanying question, whether to use self-training (ST) for the source language monolingual corpus~\cite{he2019selftraining,yarowsky-1995-unsupervised} or back-translation (BT) for the target language monolingual corpus~\cite{bojar-tamchyna-2011-improving,sennrich-etal-2016-improving,poncelas2018investigating}. We reframe the question as ``\textit{with abundant monolingual data from both languages, but limited computation resources, which data (together with the corresponding semi-supervised learning method) should we choose?}''


In previous work, BT is the most widely used technique~\cite[p. 15]{bojar-etal-2018-findings,edunov-etal-2018-understanding,ng-etal-2019-facebook,barrault-etal-2019-findings}, and is reported to outperform ST~\cite{burlot-yvon-2018-using}. 
Another line of previous work inspects the performance gain by BT. Some argue that BT is helpful mostly on the test set aligned with the model~\cite[Appendix A Table 7]{toral-etal-2018-attaining,freitag-etal-2019-ape,edunov-etal-2020-evaluation} but not the unaligned test set, while others show that BT improves performance on both test sets~\cite{edunov-etal-2020-evaluation}.

We re-inspect the two previous lines of work, and find (1) BT does \textit{not always} outperform ST, especially when ST can make use of the monolingual data in the original language of the test set (to produce pseudo-aligned training data), and (2) the performance gain by BT is \textit{not always} larger on the unaligned test set, but depends on the model direction, especially when BT generates pseudo-aligned training data with the test set.

\begin{table}[t]
    \centering
    \small
    \begin{tabular}{ll|l}
    \toprule
    \multicolumn{3}{c}{\textbf{English-to-French (en-to-fr) Translation}} \\
    & Test 1 (en, fr$^*$) & Test 2 (en$^*$, fr) \\ \hline
    Sup. on Equal Mix & 16.16 & 16.65 \\
    + ST (en, ${\text{fr}}^{**})$ & \textbf{+2.04 (Aligned)} & +1.74 \\
    + BT (${\text{en}}^{**}$, fr) & +1.91 & \textbf{+2.45 (Aligned)} \\
    \bottomrule
    \multicolumn{3}{c}{\textbf{French-to-English (fr-to-en) Translation}} \\
    & Test 1 (fr, en$^*$) & Test 2 (fr$^*$, en) \\ \hline
    Sup. on Equal Mix & 18.39 & 15.09 \\
    + ST (fr, ${\text{en}}^{**})$ & \textbf{+2.64 (Aligned)} & +2.24 \\
    + BT (${\text{fr}}^{**}$, en) & +2.17 & \textbf{+3.26 (Aligned)} \\
    \bottomrule
    \multicolumn{3}{c}{\textbf{English-to-German (en-to-de) Translation}} \\
    & Test 1 (en, de$^*$) & Test 2 (en$^*$, de) \\ \hline
    Sup. on Equal Mix & 10.59 & 8.80 \\
    + ST (en, ${\text{de}}^{**})$ & \textbf{+1.92 (Aligned)} & +1.60 \\
    + BT (${\text{en}}^{**}$, de) & +1.86 & \textbf{+2.25 (Aligned)} \\
    \bottomrule
    \multicolumn{3}{c}{\textbf{German-to-English (de-to-en) Translation}} \\
    & Test 1 (de, en$^*$) & Test 2 (de$^*$, en) \\ \hline
    Sup. on Equal Mix & 11.99 & 13.46 \\
    + ST (de, ${\text{en}}^{**})$ & \textbf{+2.28 (Aligned)} & +1.25 \\
    + BT (${\text{de}}^{**}$, en) & +1.99 & \textbf{+3.72 (Aligned)} \\
    \bottomrule
    \end{tabular}
    \caption{Performance on the en-fr and en-de test sets of \textit{newstest2014}. 
    There are two test sets for each task,
    where $^*$ marks the translated language. We use an equal mixture of supervised data in two human translation directions (``\textit{Sup.~on Equal Mix}'').
    Both ST and BT generate pseudo-parallel data (marked by $^{**}$), with which we find that \textbf{aligned} directions between the test set and the pseudo-parallel data lead to larger performance gain.}
\vspace{-0.7em}
    \label{tab:ssl_newstest}
\end{table}

We implement BT by~\citet{edunov-etal-2020-evaluation}, and ST by~\citet{he2019selftraining}. To fairly compare the performance of ST vs. BT, for each language pair $X$ and $Y$, we split half both training corpora into \forwarddata{}-Half1,  \forwarddata{}-Half2, \backwarddata{}-Half1, and \backwarddata{}-Half2. We construct the supervised training data as an equal mix (i.e., $\alpha$=50) combining \forwarddata{}-Half1 and \backwarddata{}-Half1. The development data is the combination of both development sets, which is also an equal mix. 

To train ST or BT, we use the second halves of the training data only as the monolingual corpora. 
For example, if the translation task is English-to-German translation, ST 
generates a pseudo-parallel corpus with original English paired with machine-translated pseudo-German, which we denote as (en, de$^{**}$). For readability, we mark the machine-translation direction with ST and BT by $^{**}$ and the human translation direction by $^*$.

Our hypothesis is that the machine-translated text pairs (en, de$^{**}$) will also show similar properties as the human-translated training data (en, de$^*$). Specifically, the more the pseudo-training data is aligned with the test set, the higher performance the semi-supervised learning method will achieve.
This is confirmed by the experiment results in Table~\ref{tab:ssl_newstest}, where, across all settings, no matter which semi-supervised learning method is used, when the pseudo-training data has the same translation direction as the test set, the resulting performance is generally higher. The experiments conducted on \causalmt{} test sets also generally show the same trend, and, due to the space limit, we include the results in the Appendix Table~\ref{tab:ssl}.


\section{Causal Effect of Data-Model Direction Match}
\label{sec:causal_effect_estimation}
The second contribution of this work is to inspect how much another factor, the data-model direction match, causally affects the MT performance.
Formally, our research question is that, for a given translation task \forwardmodel{}, considering an equal mix of the test set, does the human translation direction of the training data still matter? If so, how large is the effect, and is it language-/task-dependent?


This section will use causal inference to isolate the effect of data-model direction match from other possible confounders and discuss its effect in different languages and translation tasks.


\subsection{Correlation in Previous Experiments}
Our previous experiments show that data-model direction match \textit{correlates} with MT performance. 
Specifically, for each translation task in Table~\ref{tab:consis_train_test}, there is a clear difference between the causal learning and anticausal learning model. We copy this naïve difference to the ``Diff'' column of Table~\ref{tab:topic_control_bleu}.

This naïve difference represents $\mathbb{E}_{\mathrm{aligned\_corpus}}[{S}] - \mathbb{E}_{\mathrm{unaligned\_corpus}}[{S}]$, which compares the performance score ${S}$ on the aligned corpus and ${S}$ on the unaligned corpus, without controlling for potential confounders. The famous ``correlation does not imply causation'' implies that this formulation cannot answer the causal question, as the two expectation terms are taken over two different distributions that are not necessarily comparable. However, by Reichenbach's Common Cause Principle \cite{reichenbach1956direction}, correlation implies the existence of some common cause behind the two correlated variables, which motivates us to investigate the causal relationship in the next section.

\subsection{Setup of the Causal Effect Estimation}\label{sec:match_setup}
\paragraph{Formulating the causal effect.}

Instead of just correlational analyses, we aim to estimate the average causal effect (ACE) \cite{holland1988causal,pearl2009causality} of the data-model direction match (i.e., causal vs. anticausal learning) ${M}$ on the translation performance ${S}$:\footnote{Note that there are two notions of causality here, one is the intervention we are interested in, namely the data-model direction match, known as causal vs. anticausal learning, and the other is the meta-level causality we are interested in, namely how much the data-model direction match (as a binary variable) causally affect the translation performance.} 
\begin{align}
   \label{eq:ate}
   \begin{split}       
      \mathrm{ACE} & = P({S}=s | \mathrm{do}({M}=1)) \\
    & - P({S}=s | \mathrm{do}({M}=0))
   ~,
   \end{split}
\end{align}
where, according to $\mathrm{do}$-calculus~\cite{pearl1995causal} in causal inference, the operator $\mathrm{do}({M}=0 \text{ or } 1)$ means to intervene on the data-model direction match to be 0 (i.e., anticausal learning) or 1 (i.e., causal learning).
The ACE formulation is about how much the model performance ${S}$ will differ if intervening the data-model direction match ${M}$ to be 0 or 1.

To estimate the ACE, we first draw the causal graph considering all variables that can interfere with the relationship between data-model direction match and MT performance. The main additional factors we need to control for are in the causal graph in Figure~\ref{fig:causal_graph_de_en}. We make the assumption that it is very likely that the two corpora of different human translation directions also vary by sentence lengths and the distribution of content~\cite{bogoychev2019domain} due to a hidden confounder (i.e., a common cause) such as the nature of Europarl. Note that since our research question is about which training data matters for given a translation task, the data-model direction match is equivalent to the human translation direction of the training data, as the model translation direction is fixed.

\begin{figure}[t]
    \centering
    \includegraphics[width=\columnwidth]{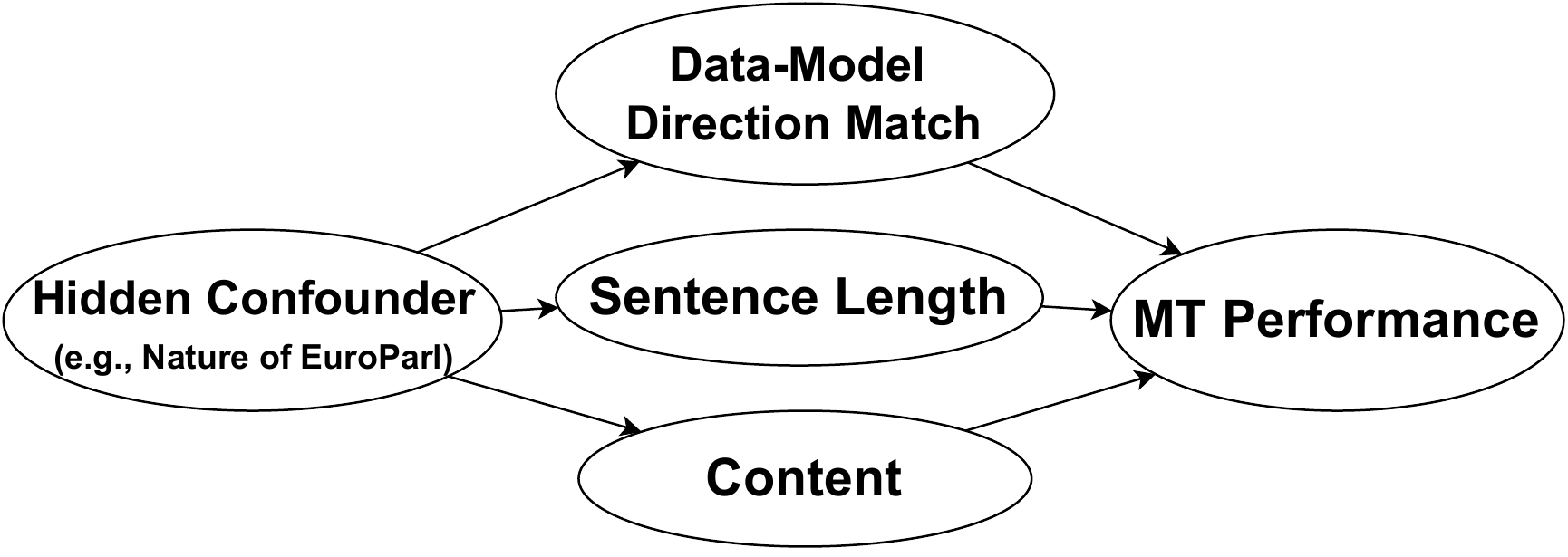}
    \caption{Causal graph about how the data-model direction match $M$ affects MT performance score $S$, considering the other covariates $Z$ including the distribution of sentence lengths and content.}
    \vspace{-0.5em}
    \label{fig:causal_graph_de_en}
\end{figure}
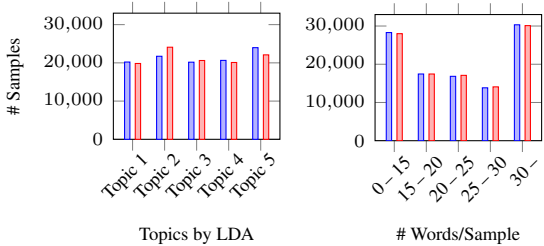
\begin{figure}[t]
\begin{subfigure}[t]{0.49\columnwidth}
    \begin{tikzpicture}  
\begin{axis}  
[  
    ybar=2pt, 
    yticklabel style={/pgf/number format/.cd,
    scaled y ticks = false,
        },
    bar width=2pt,
    legend style={at={(1,1)}, 
      anchor=north,legend columns=-1},     
    symbolic x coords={{Topic 1},{Topic 2},{Topic 3},{Topic 4},{Topic 5}},  
    x tick label style={
                rotate=45,
                anchor=east,
                xshift=0.5em,
                yshift=-0.5em,
            },
    xtick=data,  
    ymin=0,
    ymax=33000,
    enlarge x limits=0.15,
    ylabel={\# Samples},
    xlabel={Topics by LDA},
    width=\linewidth,
    ]  
\addplot coordinates {({Topic 1},20216) ({Topic 2},21725) ({Topic 3},20177) ({Topic 4},20645) ({Topic 5},23975)}; 
\addplot coordinates {({Topic 1},19838) ({Topic 2},24092) ({Topic 3},20618) ({Topic 4},20098) ({Topic 5},22092)}; 
\end{axis}  
\end{tikzpicture}
\end{subfigure}
\hfill
\begin{subfigure}[t]{0.49\columnwidth}
\begin{tikzpicture}  
\begin{axis}  
[  
    yticklabel style={/pgf/number format/.cd,
    scaled y ticks = false,
        },
    ybar=2pt, 
    bar width=2pt,
    legend style={at={(0.5,-0.5)}, 
      anchor=north,legend columns=-1},     
    symbolic x coords={0 -- 15,15 -- 20,20 -- 25,25 -- 30,30 --},  
    x tick label style={
                rotate=45,
                anchor=east,
                xshift=0.5em,
                yshift=-0.5em,
            },
    xtick=data,  
    ymin=0,
    ymax=33000,
    enlarge x limits=0.15,
    xlabel={\# Words/Sample},
    ]  
\addplot coordinates {(0 -- 15,28293) (15 -- 20,17449) (20 -- 25,16840) (25 -- 30,13836) (30 --,30321)}; 
\addplot coordinates {(0 -- 15,28001) (15 -- 20,17449) (20 -- 25,17097) (25 -- 30,14085) (30 --,30107)};
\end{axis}  
\end{tikzpicture}  
\end{subfigure}
    \caption{Distribution of topics and sentence lengths after matching, using the German-English language pair (between \textcolor{blue!70}{De\humantrans{}En} and \textcolor{red!70}{De\revhumantrans{}En}) as an example.}
    \label{fig:after_match_corpus_sent_len}
    \label{fig:after_match_corpus_topic}
    \vspace{-0.5em}
\end{figure}

\begin{table}[t]
    \centering
    \small
    \resizebox{\columnwidth}{!}{
    \begin{tabular}{p{0.11\columnwidth}p{0.8\columnwidth}}
\toprule
\textbf{Corpus} & \textbf{Matched Pairs} \\\midrule
De\humantrans{}En & Let me make some comments on specific points. \\
De\revhumantrans{}En & However, I have one or two points. \\\midrule

De\humantrans{}En & From our perspective, it is now urgently necessary that the Council also accepts this proposal, so that the negotiations can commence as soon as possible. \\
De\revhumantrans{}En & That greater urgency has been recognised in the Council suggestion that we should have an intergovernmental conference beginning next year, something which we subscribe to. \\\midrule

De\humantrans{}En & We want a European Union that is strong, effective and democratic, and all those who want to make it no more than a free trade zone within Europe will have a fight on their hands. \\
De\revhumantrans{}En & I agree that the European Union needs an integrated, coherent and consistent European energy policy that maintains Europe's competitiveness, safeguards our environmental objectives and ensures our security of supply. \\
\bottomrule
    \end{tabular}
    }
    \caption{Examples of matched pairs between the De\humantrans{}En and  De\revhumantrans{}En corpora.}
    \vspace{-0.5em}
    \label{tab:matched_examples}
\end{table}

Given the causal graph in Figure~\ref{fig:causal_graph_de_en}, the ACE in Eq.~\eqref{eq:ate} can be calculated by conditioning on the set of variables ${Z}$ which blocks the backdoor paths~\cite{pearl1995causal} between ${M}$ and ${S}$. (${Z}$ fits the backdoor criterion~\cite{pearl1993comment} in that the sentence lengths and content block all non-directed paths from ${M}$ to ${S}$, and neither is a descendant of any node on the directed path from ${M}$ to ${S}$.)
An intuitive interpretation can be that when we directly look at the correlation between the data-model direction match and MT performance, it might also be due to that different corpora have different distributions of sentence lengths and content. Therefore, we need to control the sentence lengths and content so that the performance difference will be solely due to the data-model direction match.

Formally, the ACE using the $\mathrm{do}$-notation can be calculated by conditioning on ${Z}$. Specifically, we integrate over the distribution of $P({Z})$, and calculate the difference in the conditional probability distribution $P({S}=s | {M}=1, {Z}=\bm{z}) - P({S}=s | {M}=0, {Z}=\bm{z})$ of ${S}$ given the data-model direction match value ${M}$ conditioned on the other key variables ${Z}$ for each of its possible value $\bm{z}$, as shown in Eq.~\eqref{eq:ate_prob}:
\begin{align}  
   \label{eq:ate_prob}
   \begin{split}        
    \mathrm{ACE} & = \int_{\bm{z}} [(P({S}=s | {M}=1, {Z}=\bm{z}) 
    \\
    & - P({S}=s | {M}=0, {Z}=\bm{z})) P(\bm{z}) ]
   \end{split} \\
  \begin{split}
    & = \mathbb{E}_{\bm{z}} [{S} | {M}=1, {Z}=\bm{z}]
    \\
   \label{eq:ate_expectation}
    & - \mathbb{E}_{\bm{z}} [{S} | {M}=0, {Z}=\bm{z}]
    ~.
  \end{split}
\end{align}
Finally, we estimate it by comparing the expected values of the model performance score ${S}$ given ${M}=0 \text{ or } 1$ over all possible values of the covariate ${Z}$ as in Eq.~\eqref{eq:ate_expectation}.

\begin{table*}[t]
    \centering
    \setlength\tabcolsep{4pt}
    \resizebox{\textwidth}{!}{
    \begin{tabular}{lcccc|lccccccc}
    \toprule
    \multicolumn{5}{c|}{\textbf{English-to-German (en-to-de) Translation}} & \multicolumn{5}{c}{\textbf{German-to-English (de-to-en) Translation}} \\
    & T1 (en, de$^*$) & T2 (en$^*$, de) & \textbf{ACE} & Diff &    & T1 (de, en$^*$) & T2 (de$^*$, en) & \textbf{ACE} & Diff \\ \hline
    Cau. (en, de$^*$) & 21.88 & 28.77 & \multirow{2}{*}{\textbf{+3.13}} & \multirow{2}{*}{+1.75} &     Cau. (de, en$^*$) & 31.70 & 28.68 & \multirow{2}{*}{\textbf{-1.89}} & \multirow{2}{*}{-2.14} \\
    Ant. (en$^*$, de) & 25.33 & 22.19 &  &  &     Ant. (de$^*$, en) & 26.35 & 35.92 &  &  \\

    \bottomrule
    \multicolumn{5}{c|}{\textbf{French-to-German (fr-to-de) Translation}} & \multicolumn{5}{c}{\textbf{German-to-French (de-to-fr) Translation}} \\
    & T1 (fr, de$^*$) & T2 (fr$^*$, de) & \textbf{ACE} & Diff &    & T1 (de, fr$^*$) & T2 (de$^*$, fr) & \textbf{ACE} & Diff \\ \hline
    Cau. (fr, de$^*$) & 18.36 & 25.45 & \multirow{2}{*}{\textbf{+5.57}} & \multirow{2}{*}{+5.0} &     Cau. (de, fr$^*$) & 32.25 & 29.98 & \multirow{2}{*}{\textbf{-3.58}} & \multirow{2}{*}{-2.71} \\
    Ant. (fr$^*$, de) & 20.46 & 17.78 &  &  &     Ant. (de$^*$, fr) & 28.07 & 37.74 &  &  \\

    \bottomrule
    \multicolumn{5}{c|}{\textbf{French-to-English (fr-to-en) Translation}} & \multicolumn{5}{c}{\textbf{English-to-French (en-to-fr) Translation}} \\
    & T1 (fr, en$^*$) & T2 (fr$^*$, en) & \textbf{ACE} & Diff &    & T1 (en, fr$^*$) & T2 (en$^*$, fr) & \textbf{ACE} & Diff \\ \hline
    Cau. (fr, en$^*$) & 33.77 & 37.42 & \multirow{2}{*}{\textbf{+1.43}} & \multirow{2}{*}{+2.53} &     Cau. (en, fr$^*$) & 42.60 & 37.34 & \multirow{2}{*}{\textbf{-0.14}} & \multirow{2}{*}{-0.65} \\
    Ant. (fr$^*$, en) & 37.67 & 32.09 &  &  &     Ant. (en$^*$, fr) & 38.05 & 42.03 &  &  \\

    \bottomrule
    \multicolumn{5}{c|}{\textbf{Spanish-to-English (es-to-en) Translation}} & \multicolumn{5}{c}{\textbf{English-to-Spanish (en-to-es) Translation}} \\
    & T1 (es, en$^*$) & T2 (es$^*$, en) & \textbf{ACE} & Diff &    & T1 (en, es$^*$) & T2 (en$^*$, es) & \textbf{ACE} & Diff \\ \hline
    Cau. (es, en$^*$) & 37.79 & 33.64 & \multirow{2}{*}{\textbf{+12.25}} & \multirow{2}{*}{+0.63} &     Cau. (en, es$^*$) & 39.04 & 33.68 & \multirow{2}{*}{\textbf{+3.50}} & \multirow{2}{*}{+3.79} \\
    Ant. (es$^*$, en) & 21.69 & 25.24 &  & &     Ant. (en$^*$, es) & 30.76 & 34.96 &  &  \\

    \bottomrule
    \multicolumn{5}{c|}{\textbf{French-to-Spanish (fr-to-es) Translation}} & \multicolumn{5}{c}{\textbf{Spanish-to-French (es-to-fr) Translation}} \\
    & T1 (fr, es$^*$) & T2 (fr$^*$, es) & \textbf{ACE} & Diff &    & T1 (es, fr$^*$) & T2 (es$^*$, fr) & \textbf{ACE} & Diff \\ \hline
    Cau. (fr, es$^*$) & 37.09 & 43.40 & \multirow{2}{*}{\textbf{+5.84}} & \multirow{2}{*}{+2.22} &     Cau. (es, fr$^*$) & 41.67 & 41.57 & \multirow{2}{*}{\textbf{-2.74}} & \multirow{2}{*}{-1.11} \\
    Ant. (fr$^*$, es) & 38.45 & 36.20 &  &  &     Ant. (es$^*$, fr) & 39.36 & 46.62 &  &  \\
    \bottomrule
    \end{tabular}
    }
    \caption{BLEU points of causal learning (Cau.) vs. anticausal (Ant.) directions after controlling for topics and lengths. 
    We calculate the ACE by taking each model's average performance on T1 and T2, and comparing how much causal models outperform anticausal models (\textbf{ACE}). In comparison, we show the naïve difference (Diff) by directly comparing the results of causal and anticausal models in Table~\ref{tab:consis_train_test} without topic and length control.
    }
    \vspace{-0.7em}
    \label{tab:topic_control_bleu}
\end{table*}

\myparagraph{Causal effect estimation by matching.}
To estimate the ACE in Eq.~\eqref{eq:ate_expectation}, the intuition is that we need to take care of the covariates in ${Z}$ so that the aligned setting and the unaligned setting are comparable. We follow the covariate matching method in causal inference~\cite{rosenbaum1983central,iacus2012causal} and adjustment in the high-dimensional setting of text~\cite{roberts2020adjusting,veitch2020adapting}. Specifically, matching is a method in causal inference to subsample the treated (i.e., the aligned corpus with the model direction) and control samples (i.e., the unaligned corpus with the model direction) so that the covariates of interest are matched.

We aim to match pairs of samples, one from the causal corpus and the other from the anticausal corpus, where we constrain them to share similar contents and similar sentence lengths.
In our implementation, for each sentence in the causal corpus, we select its most similar match in the anticausal corpus using Dinic's maximal matching algorithm~\cite{dinic1970algorithm}.

Empirically, we limit the sentence length ratio of each matched pair to be no larger than 1.1 and the content to have a cosine similarity larger than 0.7, following the threshold to match a content-similar pseudo-parallel corpus in \citet{jin-etal-2019-imat}. 
To calculate the content-wise similarity of a pair of samples, we represent each sentence by the sentence BERT embedding~\cite{reimers-gurevych-2019-sentence}. In cases of multiple languages as candidates to match the sentence embeddings in, we set a prioritization order of ``En$>$De$>$Fr$>$Es'' for sentence embedding matching. The prioritization order roughly follows the training data sizes of the multilingual version \cite{reimers-gurevych-2020-making} of the sentence transformer in the four languages.
Note that since the set of factors to control is in a high-dimensional vector space, it is less realistic to use other common matching methods such as propensity score stratification and matching, as pointed out by~\citet{roberts2020adjusting}.




We check the quality of our matching heuristics. Taking the German-English language pair as an example, we plot the distributions of topics by Latent Dirichlet Allocation (LDA) topic modeling~\cite{blei2001latent} and distributions of sentence lengths across the De\humantrans{}En and De\revhumantrans{}En corpora in \cref{fig:after_match_corpus_topic}. We also list some example matched pairs in English in \cref{tab:matched_examples}.
Further statistics of the matched corpora are in \cref{appd:matched_pairs}.

Finally, based on the matched datasets that control for the sentence lengths and contents, ACE can be calculated by the average difference in MT performance of models trained on the two directions of the new datasets.

\vspace{-0.5em}
\subsection{Causal Effect Results}
We have three observations from the results in \cref{tab:topic_control_bleu}: (1) The data-model direction match is a clear cause for MT performance. The ACE of data-model direction match on MT performance can be up to 12.25 BLEU points, for example, in the Spanish-to-English translation task. (2) The ACE varies by language and translation tasks. For the English-Spanish language pair, both translation directions get higher BLEU points if the models are trained in the causal learning direction. For other language pairs, the causal effects of data-model direction match are clear, although varying from positive to negative values, the reasons for which are worth future studies. (3) The results of naïve differences (Diff) are, in most cases, smaller than that of the causal analysis by ACE. This indicates that the correlational analysis neglects other important factors such as the sentence length and content, which might also be reflected in the overall difference. The causal analysis is a more appropriate method to isolate the influence of the data-model direction match.

\section{Related Work}
Linguistic studies have long observed the distinct properties of translationese from text originally authored in the same language~\cite{toury1980search,gellerstam1986translationese,baker1993corpus,toury1995descriptive}.
Recent work in MT identifies that the source side of the translationese-to-original portion of the test sets (i.e., test sets unaligned with the model direction) is easier~\cite{graham-etal-2020-statistical}, echoing with many previous observations~\cite{toral-etal-2018-attaining,lembersky-etal-2012-language,laubli-etal-2018-machine} and thus some suggest to exclude this portion from future test sets, in particular human evaluations~\cite{toral-etal-2018-attaining,zhang-toral-2019-effect,graham-etal-2020-statistical,barrault-etal-2019-findings}. Nevertheless, \newcite{freitag-etal-2019-ape} demonstrate that it is worth reporting automatic metric scores on both directions separately as both types of test sets evaluate different properties of translation quality. In follow-up work, \newcite{freitag2020bleu} introduce paraphrased test sets that overcome the problems of translationese for test sets, not solving the problem for the training data though.

Based on this speculated inflation of MT performance due to translationese in the test set, further work inspects what previous conclusions about the effectiveness of MT models should be recalibrated. Some discover that models with BT mostly improve on the inflated test set but not the other more challenging portion~\cite[Appendix A Table 7]{toral-etal-2018-attaining,freitag-etal-2019-ape, edunov-etal-2020-evaluation} and raise concerns that BT is not as effective as expected. Others argue that BT can still improve on both test sets~\cite{edunov-etal-2020-evaluation}.

While for almost all test sets the original language of each example is known, the majority of training data does not contain this meta information.
Studies of the impact of translationese on training examples are thus mostly based on Europarl where meta information is given. For instance, \newcite{ozdowska-way-2009-optimal} argue that the original side of each training sample is important when building a statistical machine translation (SMT) system.
\newcite{kurokawa-etal-2009-automatic,koppel-ordan-2011-translationese,sominsky-wintner-2019-automatic} build classifiers to automatically detect the direction of each training sample.
\newcite{riley-etal-2020-translationese} use a CNN classifier to separate the training data at scale and bias the NMT model via tagging to generate translations that look like the original text. Human evaluation demonstrates that this produces more accurate and natural translations.
Further, \newcite{amponsah-kaakyire-etal-2021-rely} investigate the impact of training samples generated with pivot (``relay'') languages. 

Our work differs from all previous work in that we conduct causal inference~\cite{pearl2009causality,peters2017elements} to contribute causal insights on how translationese affects MT.

\section{Future Work}
We list several directions for potential future work: (1) It will be meaningful to explore whether the conclusions of this paper can generalize to higher-resource data and a wider variety of languages.
(2) In real-world MT systems, it is important not only to answer whether aligned training data or unaligned training data is better (when deciding how to distribute the budget to collect data for a usage scenario with a fixed direction), but also to investigate how to utilize the mixed training data (when trying to make the best use of the existing data) and contribute to the best possible translation systems.
\citet{riley-etal-2020-translationese,larkin-etal-2021-like,freitag-etal-2022-natural} suggest adding a tag per sample specifying the direction to make use of the unaligned data. Future work can also explore if there can be an end-to-end model jointly inferring the human translation direction and signaling the model to deal with the aligned and unaligned directions differently. 
(3) In our study, \cref{tab:topic_control_bleu} mainly confirms the causal effects of the data-model direction match on the MT performance, but the actual reasons for positive and negative ACEs are still worth further investigation.



\section{Conclusion}
In conclusion, this work studies the causal effects of three important factors on MT performance: the test-model, train-test, and data-model direction match. 
We provide suggestions for future study in MT, such as using more training data in the aligned direction and paying attention to whether the nature of the translation task is causal or anticausal.

\section*{Ethical Considerations}
\myparagraph{Data Privacy and Bias:}
This research mainly focuses on translation using the Europarl~\cite{koehn-2005-europarl} corpus, which is widely adopted in the community. There are no data privacy issues or bias against certain demographics with regard to this dataset.

\myparagraph{Potential Use:}
The potential use of this study is to improve future MT practice in terms of both evaluation and training. 

\myparagraph{Generalizability:}
Most conclusions in this study are language-agnostic and potentially help MT in all language pairs, although due to the limitations of available data, the study mainly uses the common languages, English, German, French, and Spanish, in a relatively low-resource setting of around 100K to 200K data. It will be meaningful to explore whether the conclusions of this paper can generalize to higher-resource data and a wider variety of languages. There is a possibility that the findings of the study will need to be further adjusted in different settings,
which we strongly encourage future work to explore.

\myparagraph{Limitations:}
First, 
the current study mainly looks into clear cases of causal or anticausal learning, but there can potentially be a third case where both languages are translated from a third language, as pointed out in~\citet[Figure 1]{riley-etal-2020-translationese}, which is worth exploring for future work.

Second, this work extracts human translation directions from the Europarl corpus, with the assumption that the speakers at the European Parliament tend to be native speakers. It might also be possible that the Europarl corpus contains text annotated as originals but from non-native speakers, but since the European Parliament is a highly formal and important venue, the speakers tend to be at least proficient users of that language, if not native. In addition, for non-English languages in our corpus, it is highly likely that the speech comes from a native speaker.

Third, in addition to the length and content factors considered in this work, it could be interesting to look at other factors that can constitute the ${Z}$ variable in \cref{eq:ate_expectation}. Some motivations include that the data-model direction match seems to be a clear cause for the MT performance, and the fact that it does not always show a very large causal effect might mean that there are additional hidden variables to take into consideration.

Lastly, due to financial budgets, we did not use human evaluation in addition to the BLEU score reported in this work. We released all outputs of our model so future work can feel free to evaluate our results by human evaluation or various other automatic evaluation metrics. See more discussions on recommended evaluation metrics in \cref{appd:evaluation_metrics}.
\section*{Author Contributions}
\myparagraph{Jingwei Ni} conducted most of the MT experiments, collected the most updated version of \causalmt{} dataset, and performed various analyses.

\myparagraph{Zhijing Jin} designed the project, conducted the causal inference experiments, structured the first version of the codes to run the MT experiments, and collected the first version of \causalmt{} dataset which is also used in our previous study \cite{jin2021causal}.

\myparagraph{Markus Freitag} helped design the storyline of the translationese part, gave insights on what analyses are important, and provided suggestions on the evaluation.

\myparagraph{Mrinmaya Sachan} and \textbf{Bernhard Schölkopf} guided the project and provided substantial contributions to the storyline of the paper.

\myparagraph{Everyone} contributed to writing the paper.

\section*{Acknowledgments}

We sincerely thank Prof Rico Sennrich for his constructive suggestions on improving the message we present from our experiments to the MT community, and Isaac Caswell at Google Translate for proofreading our paper.
We thank Claudia Shi for the insightful discussions on the matching method and suggestions on papers to cite. We thank Shaoshu Yang and Di Jin for supporting us with computational resources. We thank Zhiheng Lyu for helping to code a fast version of the sentence similarity matching.

This material is based in part upon works supported by the German Federal Ministry of Education and Research (BMBF): Tübingen AI Center, FKZ: 01IS18039B; the Machine Learning Cluster of Excellence, EXC number 2064/1 – Project number 390727645; the John Templeton Foundation (grant \#61156); a Responsible AI grant by the Haslerstiftung; and an ETH Grant
(ETH-19 21-1).

Zhijing Jin is supported by the Open Phil AI Fellowship and the Vitalik Buterin PhD Fellowship from the Future of Life Institute.

\bibliography{ref/custom_anthology_0,ref/custom_anthology_1,ref/custom_anthology_2,ref/custom_anthology_3,ref/custom_anthology_4,custom_translation,custom_others}
\bibliographystyle{sec/acl_natbib}

\clearpage
\newpage

\appendix

\section{Reproducibility, License, and Copyright}
We open-source our codes and datasets, which are both uploaded to the submission system. In our data, we include all three variations: the full \causalmt{} dataset, the split used for the semi-supervised learning experiments, and the subset after matching the contents and sentence lengths.
In our codes, we include all commands with hyperparameters to help future work to reproduce our results.

The codes and data are under MIT license. 
Note that the Europarl dataset has no copyright restriction, according to its official website.\footnote{\url{https://www.statmt.org/europarl/}}

\section{Linguistic Property Analysis}\label{appd:linguistic}
We also open-source the codes to calculate the linguistic properties of our dataset in Table~\ref{tab:corpus_complexity}. 
We use the Python library Stanza\footnote{\url{https://stanfordnlp.github.io/stanza/}}~\cite{qi-etal-2020-stanza} to tokenize the sentences when calculating the number of sentences per sample. For computational efficiency, we use NLTK\footnote{\url{https://www.nltk.org/}}~\cite{bird2009natural} to tokenize the words and count the vocabulary.

We use the Python library spaCy\footnote{\url{https://spacy.io/}}~\cite{spacy2} to calculate the 
punctuation per sample. We use a passive voice checker\footnote{ \url{https://github.com/armsp/active_or_passive}} (only available in English). For the expansion factor, we formatted \cref{tab:corpus_complexity} using the ratio of the two languages in a descending alphabetical order of each language pair. In our table, it happens to be the ratio of the more verbose language to the less verbose language in each language pair.

\section{Implementation Details}\label{appd:implementation_details}
\subsection{Preprocessing}

To prepare the text for the models, we follow the preprocessing scripts of fairseq~\cite{ott-etal-2019-fairseq}.\footnote{\url{https://github.com/pytorch/fairseq/}}
Specifically, we use the Moses tokenizer~\cite{koehn-etal-2007-moses},\footnote{\url{https://github.com/moses-smt/mosesdecoder/blob/master/scripts/tokenizer/tokenizer.perl}} the default byte pair encoding (BPE) size of 40K subwords, and remove sentence pairs that of larger than 1.5 length ratio from the training set.

\subsection{Evaluation Script}
We use the fairseq script\footnote{\url{https://github.com/pytorch/fairseq/blob/main/fairseq_cli/generate.py}} to calculate the BLEU score~\cite{papineni-etal-2002-bleu} of each translation model, with a beam width of 5, BPE removed, and detokenized by moses.

\subsection{Model Details}
We use the sequence-to-sequence Transformer model~\cite{vaswani2017attention} implemented by the fairseq library~\cite{ott-etal-2019-fairseq}.
Specifically, we use a six-layer Transformer, a label smoothing of 0.1, a weight decay of 0.0001, a dropout of 0.3, 4000 warming updates, and a learning rate of 0.0005. All results are reported by a single run but a fixed random seed.

For the semi-supervised learning, we implement the BT model following~\citet{edunov-etal-2020-evaluation} to use the Facebook-FAIR system of the WMT'19
news shared translation task.\footnote{\url{https://github.com/pytorch/fairseq/tree/main/examples/backtranslation}} All the hyperparameters are the same as the supervised system, with a learning rate of 0.0007 on both the supervised training data and the generated pseudo-parallel corpus.
We implement the ST model by~\citet{he2019selftraining} following their script,\footnote{\url{https://github.com/jxhe/self-training-text-generation/blob/master/self_train.sh}} and also keep the hyperparameters the same as the supervised model. 

\subsection{Training Details}
We train the supervised learning model and each step in the semi-supervised learning scripts for 1000 epochs. We select the model with the best performance on the development set and report the final evaluation results on the test set.

All experiments are run on NVIDIA RTX2080 GPUs. Each supervised learning experiment takes around 32 GPU hours, and each semi-supervised learning experiment takes about 128 GPU hours.

\section{Additional Experimental Results}
\subsection{Effect of Train-Test Direction Match on Supervised Learning}
To inspect the influence of train-test direction match on the MT performance, we conduct all experiments on our 
\causalmt{} test sets and also the standard newstest2014 test sets. For the supervised learning performance, we list the performance on the \causalmt{} test sets in the main paper in Table~\ref{tab:consis_train_test}, and list the additional performance on the newstest2014 test sets in Table~\ref{tab:consis_train_test_newstest}. 
\begin{table}[t]
    \centering
    \small
    \setlength\tabcolsep{0pt}
    \begin{tabular}{cc|c|cc|ccc}
    \toprule
    \multicolumn{3}{c|}{\textbf{de-to-en Translation}} & \multicolumn{3}{c}{\textbf{en-to-de Translation}} \\
    $\alpha\%$ & T1 (de, en$^*$) & T2 (de$^*$, en) & $\alpha\%$ & T1 (en, de$^*$) & T2 (en$^*$, de) \\\hline
0\% & \adjustbox{margin=2pt,bgcolor=blue!32.38}{14.21} & \adjustbox{margin=2pt,bgcolor=red!70.00}{19.10} & 0\% & \adjustbox{margin=2pt,bgcolor=blue!9.08}{11.18} & \adjustbox{margin=2pt,bgcolor=red!42.23}{15.49} \\
25\% & \adjustbox{margin=2pt,bgcolor=blue!43.92}{15.71} & \adjustbox{margin=2pt,bgcolor=red!66.85}{18.69} & 25\% & \adjustbox{margin=2pt,bgcolor=blue!20.69}{12.69} & \adjustbox{margin=2pt,bgcolor=red!33.00}{14.29} \\
50\% & \adjustbox{margin=2pt,bgcolor=blue!52.08}{16.77} & \adjustbox{margin=2pt,bgcolor=red!62.85}{18.17} & 50\% & \adjustbox{margin=2pt,bgcolor=blue!25.38}{13.30} & \adjustbox{margin=2pt,bgcolor=red!33.31}{14.33} \\
75\% & \adjustbox{margin=2pt,bgcolor=blue!53.15}{16.91} & \adjustbox{margin=2pt,bgcolor=red!48.23}{16.27} & 75\% & \adjustbox{margin=2pt,bgcolor=blue!26.00}{13.38} & \adjustbox{margin=2pt,bgcolor=red!24.31}{13.16} \\
100\% & \adjustbox{margin=2pt,bgcolor=blue!46.31}{16.02} & \adjustbox{margin=2pt,bgcolor=red!22.38}{12.91} & 100\% & \adjustbox{margin=2pt,bgcolor=blue!25.23}{13.28} & \adjustbox{margin=2pt,bgcolor=red!5.23}{10.68} \\

    \bottomrule
    \multicolumn{3}{c|}{\textbf{en-to-fr Translation}} & \multicolumn{3}{c}{\textbf{fr-to-en Translation}} \\
    $\alpha\%$ & T1 (en, fr$^*$) & T2 (en$^*$, fr) & $\alpha\%$ & T1 (fr, en$^*$) & T2 (fr$^*$, en) \\\hline
0\% & \lightbluebox{16.61} & \lightredbox{21.33} & 0\% & \lightbluebox{16.34} & \lightredbox{23.26} \\
25\% & \lightbluebox{18.56} & \lightredbox{20.95} & 25\% & \lightbluebox{18.81} & \lightredbox{23.31} \\
50\% & \lightbluebox{20.45} & \lightredbox{21.66} & 50\% & \lightbluebox{19.75} & \lightredbox{23.20} \\
75\% & \lightbluebox{21.19} & \lightredbox{21.05} & 75\% & \lightbluebox{21.09} & \lightredbox{22.01} \\
100\% & \lightbluebox{21.43} & \lightredbox{19.30} & 100\% & \lightbluebox{20.02} & \lightredbox{19.78} \\

    \bottomrule
    \end{tabular}
    \caption{Effect of train-test direction match on the en-fr and en-de test sets of \textit{newstest2014}.}
    \label{tab:consis_train_test_newstest}
\end{table}

For better visualization of the trends, we also provide line plots of the same experimental results in Table~\ref{tab:consis_train_test}. Specifically, we plot the results of German-English translation in Figure~\ref{fig:consis_train_test_de_en} using our previous experiment results in Table~\ref{tab:consis_train_test}.
We also include the diagram of all five language pairs in Figure~\ref{fig:consis_train_test}.

In Figure~\ref{fig:consis_train_test_de_en}, we use lines with the same darkness of color for the same model trained on different data directions. Results show that the data-model direction match matters significantly. Taking the German-to-English translation models (\textcolor{orange}{- - -} and \textcolor{orange}{---}), the two data directions can cause up to 4.53 difference in BLEU points.
In the current figures, we also see that the data direction with a smaller expansion factor is a better training corpus than the other one. 

We use the same line type (dashed or solid) for models trained on the same data. Using the same data, the performance of the two different directions of models cannot be compared directly because the target language is different, causing the BLEU calculation to be different.

\colorlet{lightred}{red!50!white}
\colorlet{lightred}{red!50!white}
\colorlet{lightorange}{orange!50!white}
\colorlet{lightgreen}{green!50!white}
\colorlet{lightblue}{blue!50!white}
\colorlet{lightblack}{black!50!white}
\begin{figure}[t]
\begin{subfigure}{\columnwidth}
\centering
\begin{tikzpicture}[scale=0.9]
        \pgfplotsset{
            scale only axis,
            xmin=0, xmax=100,
            xtick={0,25,50,75,100},
            legend style={at={(1,0)},anchor=south east},
        }
        \begin{axis}[
          width=0.8\columnwidth,
          height=0.6\columnwidth,
          grid=both,
          legend style={nodes={scale=0.7, transform shape},
          legend pos={north east},
          legend cell align={left}},
          ymin=15, ymax=59,
          xlabel={Mixture Rate $\alpha$ (\%). Train Set $=\alpha\%$ $X$\humantrans{}$Y$$+ (1-\alpha\%)$ $X$\revhumantrans{}$Y$.},
          ylabel=BLEU on the Test Set of $X$\humantrans{}$Y$ (\%),
          minor tick num=1,
          label style={font=\small},
          tick label style={font=\small},
          every axis plot/.append style={thick}
        ]
        \addplot[dashed,orange]
          coordinates{
            (100, 31.33)
            (75, 31.52)
            (50, 30.86)
            (25, 28.98)
            (0, 24.68)
        }; \addlegendentry{Data: $X$\humantrans{}$Y$=De\humantrans{}En. Model: De-En}
        
        \addplot[dashed,lightorange]
          coordinates{
            (100, 26.27)
            (75, 25.44)
            (50, 24.70)
            (25, 23.23)
            (0, 20.32)
        }; \addlegendentry{Data: $X$\humantrans{}$Y$=De\humantrans{}En. Model: En-De}
        
        \addplot[smooth,orange]
          coordinates{
            (0, 27.07)
            (25, 31.92)
            (50, 34.53)
            (75, 35.40)
            (100, 35.86)
        };
        \addlegendentry{Data: $X$\humantrans{}$Y$=De\revhumantrans{}En. Model: De-En}
        
        \addplot[smooth,lightorange]
          coordinates{
            (0, 21.24)
            (25, 25.60)
            (50, 27.29)
            (75, 27.82)
            (100, 28.94)
        }; \addlegendentry{Data: $X$\humantrans{}$Y$=De\revhumantrans{}En. Model: En-De}
        
        \end{axis}
        \end{tikzpicture}

\caption{Translation performance between German and English on different mixtures of training sets combining $\alpha\%$ $X$\humantrans{}$Y$ data and $(1-\alpha\%)$ $X$\revhumantrans{}$Y$ data, where $\alpha=0,25,50,75,100$. Note that there are four settings between German and English, by varying two different data origins ($X$\humantrans{}$Y$ data $=$ De\humantrans{}En or De\revhumantrans{}En) and two different translation task directions (German-to-English (De-En) translation or English-to-German (En-De) translation).}
\label{fig:consis_train_test_de_en}
\end{subfigure}
\\
\begin{subfigure}{\columnwidth}
\begin{tikzpicture}[scale=0.65]
        \pgfplotsset{
            scale only axis,
            xmin=0, xmax=100,
            xtick={0,25,50,75,100},
            legend style={at={(1,0)},anchor=south east},
        }
        \begin{axis}[
          width=0.8\columnwidth,
          height=0.6\columnwidth,
          grid=both,
          legend style={nodes={scale=0.7, transform shape},
          legend pos={outer north east},
          legend cell align={left}},
          ymin=15, ymax=49,
          xlabel={Training Set as a Mix of $\alpha$\% $X$\humantrans{}$Y$ and 1-$\alpha$\% $X$\revhumantrans{}$Y$ { } \quad\quad},
          ylabel=Performance (BLEU) on the Test Set $X$\humantrans{}$Y$,
          minor tick num=1,
          label style={font=\small},
          tick label style={font=\small},
          every axis plot/.append style={thick}
        ]
        \addplot[dashed,orange]
          coordinates{
            (100, 31.33)
            (75, 31.52)
            (50, 30.86)
            (25, 28.98)
            (0, 24.68)
        }; \addlegendentry{Data: De\humantrans{}En. M: De-En}
        
        \addplot[dashed,lightorange]
          coordinates{
            (100, 26.27)
            (75, 25.44)
            (50, 24.70)
            (25, 23.23)
            (0, 20.32)
        }; \addlegendentry{Data: De\humantrans{}En. M: En-De}
        
        \addplot[dashed,green]
          coordinates{
            (100, 30.45)
            (75, 30.11)
            (50, 28.87)
            (25, 28.60)
            (0, 24.37)
        }; \addlegendentry{Data: De\humantrans{}Fr. M: De-Fr}
        
        \addplot[dashed,lightgreen]
          coordinates{
            (100, 22.62)
            (75, 22.88)
            (50, 22.10)
            (25, 20.94)
            (0, 18.68)
        }; \addlegendentry{Data: De\humantrans{}Fr. M: Fr-De}
        
        \addplot[dashed,blue]
          coordinates{
            (100, 41.13)
            (75, 40.81)
            (50, 39.37)
            (25, 37.41)
            (0, 35.38)
        }; \addlegendentry{Data: En\humantrans{}Fr. M: En-Fr}
        
        \addplot[dashed,lightblue]
          coordinates{
            (100, 39.07)
            (75, 39.74)
            (50, 37.40)
            (25, 36.45)
            (0, 33.97)
        }; \addlegendentry{Data: En\humantrans{}Fr. M: Fr-En}
        
        \addplot[dashed,black]
          coordinates{
            (100, 39.74)
            (75, 39.00)
            (50, 38.00)
            (25, 36.64)
            (0, 31.74)
        }; \addlegendentry{Data: En\humantrans{}Es. M: En-Es}
        
        \addplot[dashed,lightblack]
          coordinates{
            (100, 40.74)
            (75, 38.69)
            (50, 37.51)
            (25, 36.61)
            (0, 33.01)
        }; \addlegendentry{Data: En\humantrans{}Es. M: Es-En}
        
         \addplot[dashed,red]
          coordinates{
            (100, 42.46)
            (75, 42.39)
            (50, 41.94)
            (25, 40.60)
            (0, 37.32)
        }; \addlegendentry{Data: Es\humantrans{}Fr. M: Es-Fr}
        
        \addplot[dashed,lightred]
          coordinates{
            (100, 41.60)
            (75, 40.64)
            (50, 39.66)
            (25, 39.03)
            (0, 37.56)
        }; \addlegendentry{Data: Es\humantrans{}Fr. M: Fr-Es}

        
        \addplot[smooth,orange]
          coordinates{
            (0, 27.07)
            (25, 31.92)
            (50, 34.53)
            (75, 35.40)
            (100, 35.86)
        };
        \addlegendentry{Data: De\revhumantrans{}En. M: De-En}
        
        \addplot[smooth,lightorange]
          coordinates{
            (0, 21.24)
            (25, 25.60)
            (50, 27.29)
            (75, 27.82)
            (100, 28.94)
        }; \addlegendentry{Data: De\revhumantrans{}En. M: En-De}
        
        \addplot[smooth,green]
          coordinates{
            (0, 27.65)
            (25, 32.42)
            (50, 34.06)
            (75, 36.21)
            (100, 36.44)
        };
        \addlegendentry{Data: Fr\humantrans{}De. M: De-Fr}
        
        \addplot[smooth,lightgreen]
          coordinates{
            (0, 18.85)
            (25, 24.30)
            (50, 25.91)
            (75, 27.41)
            (100, 27.79)
        }; \addlegendentry{Data: Fr\humantrans{}De. M: Fr-De}
        
        \addplot[smooth,blue]
          coordinates{
            (0, 36.00)
            (25, 38.95)
            (50, 39.96)
            (75, 40.98)
            (100, 42.40)
        };
        \addlegendentry{Data: Fr\humantrans{}En. M: En-Fr}
        
        \addplot[smooth,lightblue]
          coordinates{
            (0, 32.44)
            (25, 36.45)
            (50, 38.13)
            (75, 39.74)
            (100, 40.07)
        }; \addlegendentry{Data: Fr\humantrans{}En. M: Fr-En}
        
        \addplot[smooth,black]
          coordinates{
            (0, 33.88)
            (25, 36.10)
            (50, 36.83)
            (75, 37.84)
            (100, 38.09)
        };
        \addlegendentry{Data: Es\humantrans{}En. M: En-Es}
        
        \addplot[smooth,lightblack]
          coordinates{
            (0, 31.91)
            (25, 35.94)
            (50, 37.36)
            (75, 39.11)
            (100, 40.27)
        }; \addlegendentry{Data: Es\humantrans{}En. M: Es-En}
        
        \addplot[smooth,red]
          coordinates{
            (0, 40.00)
            (25, 43.88)
            (50, 45.57)
            (75, 46.43)
            (100, 46.25)
        };
        \addlegendentry{Data: Fr\humantrans{}Es. M: Es-Fr}
        
        \addplot[smooth,lightred]
          coordinates{
            (0, 39.16)
            (25, 41.81)
            (50, 43.48)
            (75, 45.13)
            (100, 45.42)
        }; \addlegendentry{Data: Fr\humantrans{}Es. M: Fr-Es}
        \end{axis}
        \end{tikzpicture}

\caption{Translation performance between all five language pairs on different mixtures of training sets combining $\alpha\%$ $X$\humantrans{}$Y$ data and $(1-\alpha\%)$ $X$\revhumantrans{}$Y$ data, where the mixture rate $\alpha=0,25,50,75,100$. }
\label{fig:consis_train_test}
\end{subfigure}
\end{figure}

\begin{table}[t]
    \centering

    \resizebox{0.8 \columnwidth}{!}{
    \begin{tabular}{ll|l}
    \toprule
    \multicolumn{3}{c}{\textbf{German-to-English (de-to-en) Translation}} \\
    & Test 1 (de, en$^*$) & Test 2 (de$^*$, en) \\ \hline
    Sup. on Equal Mix & 25.52 & 29.02 \\
    + ST (de, ${\text{en}}^{**})$ & \textbf{+3.25 (Aligned)} & +2.59 \\
    + BT (${\text{de}}^{**}$, en) & +0.97 & \textbf{+2.67 (Aligned)} \\
    \bottomrule
    \multicolumn{3}{c}{\textbf{English-to-German (en-to-de) Translation}} \\
    & Test 1 (en, de$^*$) & Test 2 (en$^*$, de) \\ \hline
    Sup. on Equal Mix & 23.76 & 21.48 \\
    + ST (en, ${\text{de}}^{**})$ & \textbf{+1.39 (Aligned)} & +1.44 \\
    + BT (${\text{en}}^{**}$, de) & -0.63 & {+0.51 (Aligned)} \\
    \bottomrule
    \multicolumn{3}{c}{\textbf{German-to-French (de-to-fr) Translation}} \\
    & Test 1 (de, fr$^*$) & Test 2 (de$^*$, fr) \\ \hline
    Sup. on Equal Mix & 25.42 & 30.10 \\
    + ST (de, ${\text{fr}}^{**})$ & \textbf{+1.79 (Aligned)} & +1.23 \\
    + BT (${\text{de}}^{**}$, fr) & +0.35 & \textbf{+1.69 (Aligned)} \\
    \bottomrule
    \multicolumn{3}{c}{\textbf{French-to-German (fr-to-de) Translation}} \\
    & Test 1 (fr, de$^*$) & Test 2 (fr$^*$, de) \\ \hline
    Sup. on Equal Mix & 21.89 & 18.60 \\
    + ST (fr, ${\text{de}}^{**})$ & \textbf{+2.46 (Aligned)} & +2.09 \\
    + BT (${\text{fr}}^{**}$, de) & +1.07 & +0.87 (Aligned) \\
    \bottomrule
    \multicolumn{3}{c}{\textbf{English-to-French (en-to-fr) Translation}} \\
    & Test 1 (en, fr$^*$) & Test 2 (en$^*$, fr) \\ \hline
    Sup. on Equal Mix & 35.64 & 36.19 \\
    + ST (en, ${\text{fr}}^{**})$ & \textbf{+2.04 (Aligned)} & +1.89 \\
    + BT (${\text{en}}^{**}$, fr) & +0.13 & {+1.33 (Aligned)} \\
    \bottomrule
    \multicolumn{3}{c}{\textbf{French-to-Englih (fr-to-en) Translation}} \\
    & Test 1 (fr, en$^*$) & Test 2 (fr$^*$, en) \\ \hline
    Sup. on Equal Mix & 34.35 & 33.75 \\
    + ST (fr, ${\text{en}}^{**})$ & \textbf{+1.76 (Aligned)} & +2.28 \\
    + BT (${\text{fr}}^{**}$, en) & +0.43 & {+1.89 (Aligned)} \\
    \bottomrule
    \multicolumn{3}{c}{\textbf{English-to-Spanish (en-to-es) Translation}} \\
    & Test 1 (en, es$^*$) & Test 2 (en$^*$, es) \\ \hline
    Sup. on Equal Mix & 33.65 & 34.01 \\
    + ST (en, ${\text{es}}^{**})$ & {-0.10 (Aligned)} & -0.75 \\
    + BT (${\text{en}}^{**}$, es) & +0.36 & \textbf{+1.04 (Aligned)} \\
    \bottomrule
    \multicolumn{3}{c}{\textbf{Spanish-to-English (es-to-en) Translation}} \\
    & Test 1 (es, en$^*$) & Test 2 (es$^*$, en) \\ \hline
    Sup. on Equal Mix & 35.00 & 33.82 \\
    + ST (es, ${\text{en}}^{**})$ & -0.46 (Aligned) & -0.41 \\
    + BT (${\text{es}}^{**}$, en) & +0.63 & \textbf{+1.77 (Aligned)} \\
    \bottomrule
    \multicolumn{3}{c}{\textbf{Spanish-to-French (es-to-fr) Translation}} \\
    & Test 1 (es, fr$^*$) & Test 2 (es$^*$, fr) \\ \hline
    Sup. on Equal Mix & 38.30 & 40.40 \\
    + ST (es, ${\text{fr}}^{**})$ & +0.58 (Aligned) & +0.83 \\
    + BT (${\text{es}}^{**}$, fr) & +1.00 & \textbf{+2.01 (Aligned)} \\
    \bottomrule
    \multicolumn{3}{c}{\textbf{French-to-Spanish (fr-to-es) Translation}} \\
    & Test 1 (fr, es$^*$) & Test 2 (fr$^*$, es) \\ \hline
    Sup. on Equal Mix & 40.61 & 38.55 \\
    + ST (fr, ${\text{es}}^{**})$ & {-0.84 (Aligned)} & -1.09 \\
    + BT (${\text{fr}}^{**}$, es) & -0.14 & \textbf{+0.12 (Aligned)} \\
    \bottomrule
    \end{tabular}
    }
    \caption{Performance analogous to Table~\ref{tab:ssl_newstest} but on our \causalmt{} test sets.}
    \label{tab:ssl}
\end{table}




\subsection{Effect of Train-Test Direction Match on Semi-Supervised Learning}
For the semi-supervised learning performance, we show the performance on the newstest2014 test sets in Table~\ref{tab:ssl_newstest} in the main paper, and performance on the test sets of \causalmt{} in Table~\ref{tab:ssl}. Note that the decrease of ST performance on En-Es and Es-Fr pairs is possible because ST is more sensitive to the quality of the model learned on the supervised data, and these language pairs have a smaller training data size of 90K compared with 200K+ data for all the other language pairs.

\subsection{Non-BLEU Evaluation Metrics}\label{appd:evaluation_metrics}

Due to financial constraints, we did not use human evaluation in addition to the BLEU score in our study. However, we encourage future studies in this line of work to include human evaluation results, as evaluation is important given the nature of such work, and human evaluation is reported to be more reflective of the real translation quality~\cite{edunov-etal-2020-evaluation}.

To make it convenient for follow-up work, we open-source all the outputs generated by our model to our GitHub. We also encourage future work to adopt more evaluation metrics such as COMET and BLUERT, which are among the metrics that correlates the best with human judgements
according to the metric task at WMT \citet{freitag-etal-2021-results}.
COMET, for instance, correlates better with human judgements than BLEU \cite{kocmi-etal-2021-ship} and goes further string matching.

\section{Implementation Details for Causal Inference}


\subsection{Statistics of the Matched Corpora} \label{appd:matched_pairs}

We list the statistics of the matched corpora in Table~\ref{tab:matched_data_stats}, and analyze its linguistic properties in Table~\ref{tab:corpus_complexity_aligned}.
\begin{table}[t]
    \centering
    \small
    \begin{tabular}{cccc}
    \toprule
    \textbf{Human Trans. Dir.} & \textbf{Train} & \textbf{Dev} & \textbf{Test} \\ \hline
    De \humantrans En & 107K & 1K & 2K \\ 
    En \humantrans De & 107K & 1K & 2K \\
    De \humantrans Fr & 133K & 1K & 2K \\ 
    Fr \humantrans De & 133K & 1K & 2K \\
    En \humantrans Fr & 87K & 1K & 2K \\ 
    Fr \humantrans En & 87K & 1K & 2K \\
    En \humantrans Es & 47K & 1K & 2K \\ 
    Es \humantrans En & 47K & 1K & 2K \\
    Es \humantrans Fr & 50K & 1K & 2K \\ 
    Fr \humantrans Es & 50K & 1K & 2K \\
    \bottomrule
    \end{tabular}
    \caption{Dataset statistics for five language pairs after matching. Each language pair has data from two human translation directions (Human Trans. Dir.), e.g., De\humantrans{}En and De\revhumantrans{}En.}
    \label{tab:matched_data_stats}
\end{table}
\begin{table*}[t]
    \centering
    \small
    
    \setlength\tabcolsep{2pt}
    \setlength\extrarowheight{2pt}
    \resizebox{\textwidth}{!}{
    \begin{tabular}{lcc|cc|cc|cc|cc}
    
    \toprule
    & 
    De\humantrans{}En & De\revhumantrans{}En & De\humantrans{}Fr & De\revhumantrans{}Fr & En\humantrans{}Fr & En\revhumantrans{}Fr & En\humantrans{}Es & En\revhumantrans{}Es & Es\humantrans{}Fr & Es\revhumantrans{}Fr \\ \hline
    \# Words/Sample & 21.05/23.82 & 22.64/23.82 & 23.81/30.09 & 24.39/29.19 & 25.40/29.76 & 25.33/28.02 & 26.69/28.18 & 26.96/27.14 & 30.59/34.33 & 30.39/33.45 \\
     \# Sents/Sample & 1.032/1.031 & 1.041/1.020 & 1.041/1.925 & 1.056/1.902 & 1.033/1.949 & 1.040/1.872 & 1.028/1.053 & 1.057/1.070 & 1.076/2.100 & 1.068/2.088 \\
     {Sent Expansion Factor} & en:de=1.00 & en:de=0.98 & fr:de=1.85 & fr:de=1.80 & fr:en=1.88 & fr:en=1.80 & es:en=1.02 & es:en=1.01 & fr:es=1.95 & fr:es=1.96 \\
    Passive Voice (\%) & -/0.1128 & -/0.1036 & -/- & -/- & 0.1073/- & 0.1185/- & 0.1155/- & 0.1256/- & -/- & -/- \\
    \# Punctuation/Sample & 3.04/2.83 & 3.04/2.63 & 3.45/6.04 & 3.35/6.43 & 2.82/5.89 & 3.11/6.16 & 2.93/2.71 & 3.12/3.07 & 3.42/7.02 & 3.50/7.44 \\
     \# Syllables/Word & 2.002/1.744 & 2.059/1.755 & 1.988/1.553 & 2.068/1.546 & 1.758/1.562 & 1.780/1.550 & 1.760/2.022 & 1.780/2.010 & 2.010/1.567 & 2.030/1.544 \\
    Flesch Reading Ease & 31.90/35.22 & 29.30/33.78 & 35.25/46.30 & 28.0/46.1 & 31.93/45.80 & 31.09/49.91 & 30.55/50.22 & 30.05/51.94 & 48.82/43.26 & 46.87/42.04 \\
    MATTR & 58.93/52.68 & 60.58/53.31 & 59.32/52.77 & 61.74/52.89 & 53.19/52.55 & 53.38/52.32 & 53.90/54.91 & 52.33/53.90 & 53.60/51.85 & 54.84/52.29 \\
    Lexical Density & 49.15/49.24 & 50.75/49.91 & 48.86/55.30 & 50.82/55.21 & 49.99/55.18 & 50.28/55.16 & 50.24/49.76 & 49.56/48.88 & 48.72/55.02 & 50.01/55.14 \\
    {Vocab Size} & 58K/22K & 56K/23K & 78K/37K & 71K/39K & 22K/31K & 21K/31K & 19K/31K & 16K/29K & 32K/26K & 34K/29K \\
    \bottomrule
    \end{tabular}
    }
    \caption{
    Detailed characteristics of the matched dataset.
    }
    \label{tab:corpus_complexity_aligned}
\end{table*}

\subsection{Confirming the Causal Graph by Causal Discovery}

To check our causal graph assumption, we first verify whether data-model direction match is a cause for MT performance using causal discovery.

We use the causal discovery algorithm, fast causal inference (FCI)~\cite{spirtes2000constructing}, to verify that the data-model direction match causally affects the translation performance, conditioned on other factors such as the sentence length and topics.

FCI is the most appropriate causal inference method for this analysis since there might exist hidden confounders that affect the MT performance, which normal causal discovery methods such as score-based methods~\cite{heckerman1999bayesian,huang2018generalized} and other constraint-based algorithms like Peter-Clark (PC) algorithm~\cite[\S 5.4.2, pp.~84–88]{spirtes2000causation} cannot handle~\cite{clark2019review}.
FCI gives asymptotically
correct results in the presence of confounders, and outputs Markov equivalence classes, i.e., a set of
causal structures satisfying the same conditional independences.

Given a language pair $X$ and $Y$, we generate eight sets of experiment results, by varying the two training directions, two test directions, and two model directions. We extract the test samples of all eight experiments, and since each test set is 2K, there are 16K samples in total. On the 16K samples, besides keeping the label of their data-model direction match, translation performance in BLEU, we also calculate the other factors such as the test-model direction match, train-test direction match, source sentence length, and the topic vector by topic modeling on all the training data of the language pair $X$ and $Y$.
We run the FCI algorithm using the causal-learn Python package\footnote{\url{https://github.com/cmu-phil/causal-learn}} over all the variables of interest.
The implementation details are in the Appendix.

The resulting causal graph on the German-English language pair is in Figure~\ref{fig:causal_graph_de_en}. The results confirm our hypothesis that the data-model direction match (causal vs. anticausal direction) does have a causal effect on the BLEU score, together with other factors such as the sentence length and topics.

\end{document}